\RequirePackage{fix-cm}
\documentclass[doublespace]{svjour3}                     % onecolumn (standard format)
\smartqed  % flush right qed marks, e.g. at end of proof
\usepackage{graphicx}
\usepackage{subfigure}
\usepackage{graphics,epsfig, amsmath,color}
\usepackage{mathptmx}      % use Times fonts if available on your TeX system
\usepackage{natbib}
\usepackage{amsmath}  
\usepackage{caption}
\usepackage{tabularx}
\usepackage[colorinlistoftodos]{todonotes}
\journalname{Journal of Classification }
\begin{document}

\title{Active learning for binary classification with variable selection %\thanks{Grants or other notes
}
\subtitle{Active learning with variable selection}

\author{Nov. 28, 2017 }
\author{Zhanfeng Wang \and
		Yumi Kwon  \and
		Yuan-chin Ivan Chang 
		%\\ Nov. 28, 2017   %<<<<<<<<<<<<<<<<<<<<<<<<<<<<<<-----------------%etc.
}

\authorrunning{ZF Wang, YM Kwon, YCI Chang} % if too long for running head

\institute{Yuan-chin Ivan Chang \at
            128 Academia Road Section 2, Taipei, Taiwan 11529 \\
              Tel.: +886 2 27871950\\
             Fax: + 882 2 27831523\\
               \email{ycchang@sinica.edu.tw}           %  \\
         \emph{Present address:} of F. Author  %  if needed
           \and
           Zhanfeng Wang \at
              University of Science Technology of China, Hefei, 230026, China \\
           \and
          YuMi Kwon \at
             Department of Applied Statistics, Korea University
}

\date{Received: date / Accepted: date}
%\newpage

\maketitle

%\newpage

\begin{abstract}
Modern computing and communication technologies can make data collection procedures very efficient.
However, our ability to analyze large data sets and/or to extract information out from them
is hard-pressed to keep up with our capacities for data collection.
Among these huge data sets, some of them are not collected for any particular research purpose.
For a classification problem, this means that the essential label information may not be readily obtainable, in the data set in hands,
and an extra labeling procedure is required such that we can have enough label information to be used for constructing a classification model.
When the size of a data set is huge, to label each subject in it will cost a lot in both capital and time.
Thus, it is an important issue to decide which subjects should be labeled first in order to efficiently reduce the training cost/time.
Active learning method is a promising outlet for this situation, because with the active learning ideas, we can select the unlabeled subjects sequentially without knowing their label information.  
In addition, there will be no confirmed information about the essential variables for constructing an efficient classification rule.
Thus, how to merge a variable selection scheme with an active learning procedure is of interest.
In this paper, we propose a procedure for building binary classification models when the complete label information is not available in the beginning of the training stage.  We study an model-based active learning procedure with sequential variable selection schemes, and discuss the results of the proposed procedure from both theoretical and numerical aspects.
\end{abstract}

%\iffalse
% letter to editor
%We propose a procedure for building binary classification models when the complete label information is not available in the beginning of the training stage. We study using a logistic-model based active learning procedure which can simultaneously select the most informative subjects for training and find the effective variables for this classification model.  We present both theoretical proofs of properties of the proposed method and numerical results obtained from applying our methods to two real data sets.  Our method is useful and efficient when  we want to build a classification/prediction model using an existent data set while there is a lack of pre-labeled samples in this data set.  We think our results will be attractive to the readers of the Classification.
%\fi

\section{Introduction}
Classification is a common task in many research areas, and a classification rule is conventionally built under a training/testing framework, where the label information is essential in its training stage.  
Modern computing and communication technologies can make data collection procedures very efficient,
and these collected huge data sets  are not for any particular research purpose.
For classification problems, this means that the essential label information for a particular problem of interest cannot be readily obtainable in a data set for training a classification model of interest.  In this situation,  we need to conduct an extra labeling procedure such that we can have enough labeled subjects for constructing such a classification model.
Besides the size of training set, the impacts or information of subjects for building a classification model are usually not the same.
Therefore, considering the labeling cost in time and capital, how to increase the size of labeled subjects in training set via selecting the most ``informative'' subjects to be examined and labeled first in order to accelerate the learning process is an important issue. 
\cite{Deng2009} gave us a typical example of this scenario, in which a bank customer data set is used to build a money laundering prediction model. 
They treated it as a binary classification problem, and their learning subjects are then adaptively, sequentially selected without their true label information. This kind of procedures are called active learning in the machine learning literature, and there are all kinds of active learning procedures proposed targeting at different learning goals, e.g. \citet{Lewis1994, Osugi2005, Lughofer2012, Rubens2016} and references there. 

Because active learning procedures sequentially recruit new observations joining the training set to continuously ameliorate the  performance, they are naturally sequential procedures from a statistical viewpoint.  
If those new subjects are recruited according to the information obtained from the previous training subjects, then this is an adaptive sequential procedure
as in stochastic regression models, and in this situation, the observations may be no longer independent.
In the literature, depending on the learning target and  model/method used, many different observation-recruiting procedures are studied \citep{ Cohn1996, Zhu2003, Bouneffouf2016}.
Here we consider a binary-classification problem, under the active learning scenario described above, with a logistic model with adaptively selected observations.  
To collect new observations is essential in both active learning procedures and sequential experimental design methods.
A major difference between them is that in active learning applications, we already have a bunch of unlabeled samples available, and it is unlikely to observe/collect new observations at those design points based on theories of experimental design.
Instead, we will just use the information obtained from the experimental design to find the suitable subjects from the studied data set.
In addition to subject selection, we will select the effective variables under the proposed active learning setup.

Because we do not want to label and use whole data set for training, the size of learning observations is of interest in addition to the selecting scheme for active learning procedures.
Here, the proposed procedure has equipped with a data-dependent stopping criterion such that it can have a satisfactory performance when we stop the learning procedure.
The receiver operating characteristic (ROC) curve and its related indexes are popular classification performance measures, in this study we adopt the area under ROC curve (AUC)  to measure classification performances,
and study that whether our procedure has a satisfactory AUC under the proposed stopping criterion.
For active learning procedures aiming at different performance measures, 
please refer to \citet{ Cohn1996, Zhu2003, Bouneffouf2016} and references there. %

In the rest of this paper, we will first introduce the adaptive shrinkage estimate (ASE) for logistic models under adaptive design case \citep{wangchang13, Lu2015}, and use the ideas there to sequentially select the effective variables.
Based on this estimate, we study an adaptive subject selection method together with the proposed active learning procedure,
and then conduct some numerical studies to illustrate the proposed method.
We also apply our method to several real data sets, and the results are also presented, and a summary is given after that.  
Some theoretical proofs and technical details are then given as Appendix.

\section{Method}%Locate effective variables}
Under an active learning scenario described above, we want to build a binary classification rule when there is only limited labeled subjects to beginning with, and a bunch of unlabeled data, which we can used to enlarge the training set after labeling them.
In our procedure, we will simultaneously select the most informative subjects for training and find the effective variables for this classification model.
We will adopt the idea of the ASE in \cite{wangchang13} to find the effective variables during the course of an active learning procedure, where they studied a sequential estimation of linear models  with the variable selection feature, and later
\citep{Lu2015} extended their results further for generalized linear model cases.
However, there is no specific observation/subject recruiting procedures discussed in their papers.
For searching suitable observations at the training stage from the existent unlabeled data set, 
we use the D-optimality from experimental design theory as a guideline and using the method of the uncertainty sampling 
for boosting the computational efficiency.
To fix notations, we will first briefly summarize the ASE of a logistic model under adaptive design cases and
then  discuss the details of the proposed active learning procedure with these notations.

\subsection{Variable selection and adaptive shrinkage estimate}
Suppose that  $\{(y_i,x_i), i=1, 2, \cdots\}$ are random observations that satisfy $E[y_i | {\cal{F}}_{i-1}]=\mu(x_i^T\beta)$ with $e_i=y_i - E[y_i | {\cal{F}}_{i-1}]$, where ${\cal{F}}_n=\sigma\{(x_j, y_j): j=1, \ldots, n\}$ for $n \geq 1$ and ${\cal F}_0=\sigma\{\emptyset, \Omega \}$.
Hence, for each $i$, $x_i$ is ${\cal{F}}_{i-1}$-measurable as in a stochastic regression model \citep{lai-wei82}.
Suppose that $y_i, i=1, 2, \cdots$, are binary response variables satisfying a logistic model 
\begin{eqnarray}\label{model}
P(y_i=1|x_i)=\mu(x_i^T\beta)=\frac{\exp{(x_i^T\beta)}}{1+\exp{(x_i^T\beta)}},
\end{eqnarray}
where $\beta \in R^p$ is a vector of (unknown) regression parameters, and $x_i, i=1, 2, \cdots$, are covariate vector with length $p$.
Let  $\tilde{\beta}_n$ be a solution to the estimating equation:
\begin{eqnarray}\label{gle}
ln(\tilde\beta_n)\equiv\sum_{i=1}^n[y_i-
\mu(x_i^T\tilde\beta_n)]x_i=0,
\end{eqnarray}
That is, $\tilde{\beta}_n$ is the maximum likelihood estimate (MLE) for $\beta$ in (\ref{model}) with a sample of size $n$.
It follows that under some moment conditions on the covariate vectors $x_i$s, it was shown that with probability one,
%\begin{align}\label{slse1}
$\|\tilde{\beta}_n-\beta_0\|\rightarrow 0 \hbox{ as } n\rightarrow \infty$
%\end{align}
(see, for example, Chen et al, 1999).
Moreover, it is shown that as $n$ goes to infinity, $$\tilde\Sigma_n^{1/2}(\tilde{\beta}_n-\beta_0) \rightarrow  N(0, I_p)\hbox{ in distribution},$$ where $\tilde\Sigma_n=\Sigma(\tilde{\beta}_n)=\sum_{i=1}^{n}\dot\mu(x_i^T\tilde{\beta}_n)x_ix_i^{T}$, and $I_p$ is the $p\times p$ identity matrix.

Let $\nu_{max}(n)$ and
$\nu_{min}(n)$ be the maximum and minimum eigenvalues of $\sum_{i=1}^{n}\dot\mu(x_i^T\beta)x_ix_i^{T}$, respectively,
and set $L=L_n=\nu_{min}(n)/\log(\nu_{max}(n))$.
Let $\lambda\equiv\lambda(n)$ is a non-random function of $n$ and $\lambda_j=\lambda |\tilde{\beta}_{nj}|^{-\gamma}$, for each $j$, with a constant $\gamma >0$ %satisfying (\ref{cond1})
such that for some $ 0<\delta<1/2 $,
\begin{align}\label{cond1}
L^{1/2}\lambda \rightarrow 0 \text{ and }
L^{1/2+\gamma\delta}\lambda\longrightarrow \infty, \mbox{ as }
n\rightarrow \infty.
\end{align}
Then by the asymptotic properties of  $\tilde \beta_n$, we have that with probability one, as $n\rightarrow \infty$, 
\begin{align}\label{key1}
L^{1/2} \lambda_j\longrightarrow 0 \times I(\beta_{0j}\neq 0)+
\infty \times I(\beta_{0j}= 0),
\end{align}
where $I(\cdot)$ is the indicator function. (Note that here we set $\infty \times 0 = 0$.)
Thus, for a given $\epsilon > 0$,  Equation \eqref{key1} suggests that we can use the indicator
$I_{nj}(\epsilon)=I(L^{1/2} \lambda_j<\epsilon) $
to determine whether the jth component, $ 1 \leq j \leq p$, of $\tilde{\beta}_{n}$ is significantly apart from zero by a positive constant $\epsilon$.

Suppose that the following Conditions (A1) and (A2): 
\begin{itemize}
\item[] {\em (A1)}
The random error $\{e_n\}$ is a martingale
difference sequence with respect to an increasing sequence of
$\sigma$-fields $\{{\cal{F}}_n\}$ with
\begin{align}\label{sc1}
\sup_nE(|e_n|^\alpha |{\cal{F}}_{n-1})<\infty \mbox{ almost surely for some
}\alpha>2.
\end{align}
\item[] {\em (A2)}
The eigenvalues, $\nu_{max}(n)$ and
$\nu_{min}(n)$, satisfy,
with probability one, that
\begin{align}\label{sc2}
\nu_{min}(n)\rightarrow \infty \text{ and }
\log(\nu_{max}(n))=o(\nu_{min}(n)). % \mbox{ almost surely },
\end{align}
\end{itemize}
are satisfied, then we have the following theorem.
\begin{theorem}\label{Th1.1}
Suppose that observations $\{(x_{i}, y_{i}), i=1,2 \ldots\}$
satisfy Assumptions (A1) and (A2). Then for {any small $\epsilon>0$},
%\begin{align}
$I_{nj}(\epsilon)\rightarrow I(\beta_{0j}\neq 0)$  almost surely as $ n\rightarrow \infty$.  %\nonumber
%\end{align}
%\textcolor{red}
In addition that {$\hat{p}_0 \equiv \sum_{i=1}^{p}I_{nj}(\epsilon)$ is a strongly
consistent estimate of $p_0$ and $\lim_{n\rightarrow\infty} E(\hat p_0) = p_0$.}
\end{theorem}
(The proof of it is given in Appendix.)

From Theorem \ref{Th1.1}, we know that
\begin{align}\label{Irate}
L^{1/2} |I_{nj}(\epsilon)-I(\beta_{0j}\neq 0)|\longrightarrow 0 
\text{ almost surely as }n\rightarrow \infty.
\end{align}
%It is equivalent to say that for a given constant $\epsilon >0$, we are able to identify those ``effective'' variables with probability one. 
Following this result, we define an ASE $\hat\beta_{n}$ based on $\tilde \beta_{n}$:
\begin{align}\label{ASE-beta}
	\hat{\beta}_n=I_n(\epsilon)\tilde{\beta}_n,
\end{align} % to be the ASE of $\beta_0$,
where $ I_n(\epsilon) =\text{diag}\{I_{n1}(\epsilon),\cdots,I_{np}(\epsilon)\}$ is a $p\times p$ diagonal matrix.
%
%Equation \eqref{ASE-beta} defines
Then by Theorem \ref{Th1.1}, for each $j$, the $j$th component of $\tilde\beta_{n}$,  $\tilde\beta_{nj}$, is shrunk to $0$ if $I_{nj}(\epsilon)=0$, otherwise it remains unchanged.
%The $I_{nj}(\epsilon)$ depends on $\tilde \beta_n$, the components of  $\tilde \beta_{n}$ are therefore shrunk randomly and depending on the observations up to the current stage.
Thus, the estimate $\hat\beta_{n}$ is a ``shrunk'' version of $\tilde \beta_{n}$, and we will refer to it as an asymptotic shrinkage estimate (ASE) of $\beta_0$.
%Assume further that Assumption {\em (A3)} holds.

If the following Condition (A3):
\begin{itemize}
\item[]  {\em (A3)} There exists a non-random positive definite
symmetric matrix $B_n$ and a continuously increasing function
$\rho(\cdot)$ such that
\begin{align}\label{sc3}
&B_n^{-1}(\sum_{i=1}^{n}\dot\mu(x_i^T\beta)x_ix_i^{T})^{1/2}\rightarrow I_p,  \\
&\max_{1\leq i\leq n}||B_n^{-1}\dot\mu(x_i^T\beta)x_i|| \rightarrow 0 \mbox{ in probability}, \mbox{ and } \nonumber\\
&\sum_{i=1}^{n}\dot\mu(x_i^T\beta)x_ix_i^{T}/\rho(n)\rightarrow \Sigma \mbox{ almost surely}, \nonumber
\end{align}
{ where $\Sigma$ is a positive definite matrix.} 
\end{itemize}
is satisfied, then we have the asymptotic normality of $\hat\beta_n$ below.
(The proof of it will also be presented in Appendix.)
%
%Under the assumptions of Theorem \ref{Th1.1} and Assumption {\em (A3)},  we also have the following theorem.
\begin{theorem}\label{Th1.2}
Suppose the assumptions of Theorem \ref{Th1.1} are satisfied.
%\textcolor{red}
Then with probability one, (i) $\hat\beta_n \rightarrow \beta_0$  and (ii)
$\|\hat{\beta}_n-\beta_0\|=O(L^{-1/2})$ as $n \rightarrow \infty$.
(iii) If, in addition, Assumption (A3) is satisfied, then  for {any small $\epsilon >0$},
\begin{align}%\label{sasynorm}
\rho(n)^{1/2}(\hat{\beta}_n-\beta_0)\rightarrow
N(0,{I_0}\Sigma^{-1} {I_0}) ~~\text{\rm in distribution as } n\rightarrow \infty,\nonumber
\end{align}
where  ${I_0} = \mbox{diag}\{I(\beta_{01}\neq
0),\cdots,I(\beta_{0p}\neq 0)\}$ is a $p\times p$ diagonal matrix.
\end{theorem}
Condition (A3) is a regularity condition for the random design matrix.  
Theorem \ref{Th1.2} (i) and (ii) mean that $\hat\beta_n$ is a strongly consistent estimate of $\beta_0$ with a convergence rate approximately equal $L^{-1/2}$.
Theorem \ref{Th1.2} (iii) indicates that if $\beta_{0j}=0$ for some $j$, then the limiting distribution of $\hat\beta_{nj}$ will eventually degenerate to 0 when $n$ is large.
Moreover, if $\beta_{0j}\neq 0$ with some $j$, then $\hat\beta_{nj}$ and $\tilde\beta_{nj}$ has the same asymptotic distribution.

%{\color{blue}
%Applying Theorem \ref{Th1.2}, it implies that with probability one,
%$\hat\beta_n \rightarrow \beta_0$  and
%$$\|\hat{\beta}_n-\beta_0\|=O(\rho(n)^{-1/2}) \text{ as } n \rightarrow \infty.$$
%Moreover, for {any small $\epsilon >0$},
%\begin{align}\label{sasynorm}
%\rho(n)^{1/2}(\hat{\beta}_n-\beta_0)\rightarrow
%N(0,\sigma^{2}{I_0}\Sigma^{-1} {I_0}) ~~\text{\rm in distribution
%as } n\rightarrow \infty,
%\end{align}
%where  ${I_0} = \mbox{diag}\{I(\beta_{01}\neq
%0),\cdots,I(\beta_{0p}\neq 0)\}$ is a $p\times p$ diagonal matrix as defined before.
%}

\color{black}
Note that because we use a data-driven stopping criterion, the size of observations  is a random variable.  
Applying the previous theorems, we have Corollary \ref{Th1.3}.
\begin{corollary}\label{Th1.3}
Let $N(t)$ be a positive integer-valued random variable such that
$N(t)/t$ converges to 1 in probability as $t\rightarrow \infty$. If
the conditions of Theorem \ref{Th1.2} are satisfied, then $\hat{\beta}_{N(t)} \rightarrow \beta_0$ with probability one, and as $t\rightarrow \infty$
\begin{align}\label{Tasynorm1}
\rho(N(t))^{1/2}(\hat{\beta}_{N(t)}-\beta_0)\rightarrow
N(0,{I_0}\Sigma^{-1} {I_0}) ~~\text{\rm in distribution}.
\end{align}
\end{corollary}
%It says that the asymptotic properties remain with a data-driven stopping criterion.
Corollary \ref{Th1.3} states that under sequential sampling strategy if a random sample size satisfying the above assumption, then
the asymptotic distribution of ASE remains.
Based on this property, we propose a stopping criterion such that the estimate $\hat\beta_n$  satisfies
a pre-required precision when we stop to recruit new subjects for training.
A brief proof of this corollary is given below.
\paragraph{Proof of Corollary \ref{Th1.3}:}
Because the integer-valued random variable $N(t)$ satisfies that $N(t)/t$ converges to 1 in probability as $t\rightarrow \infty$, and we know that  $\hat{\beta}_{n} \rightarrow \beta_0$ with probability one as $n\rightarrow \infty$ from Theorem \ref{Th1.2},
it implies that  $\hat{\beta}_{N(t)} \rightarrow \beta_0$ with probability one.
To prove the asymptotic distribution of $\hat\beta_{N(t)}$ is sufficient to show that
$\{\rho(n)^{1/2}(\hat\beta_n-\beta_0), n=1,2,\cdots\}$ is uniform continuity in probability (ucip) \citep{Woodroofe:1982},
and in the current problem the proof will follow the arguments of \cite{Anscombe1952} \citep[see also][]{Woodroofe:1982, wangchang13} and using the H\'{a}jek-R\'{e}nyi Inequality for martingale differences \citep[see][page 247]{ChowTeicher1988}.
Hence, we omit the detailed arguments here.

%%%%
\color{black}

\section{Subject selection, variable determination and stopping criterion}\label{sec:seq-sampling}
Suppose that we already have $n$ subjects as a training set in the $n$th stage, and 
let $Y_n=(y_1, y_2, \cdots, y_n)^{T}$ be a $n \times 1$ vector of the labels of these $n$ subjects, 
and $X_n=(x_1, x_2, \cdots, x_n)$ be the corresponding $p \times n$ matrices of covariates.
Without loss of generality,
we can rearrange the components of $\hat\beta_n$ as $(\hat\beta^{T}_{n1},\hat\beta^{T}_{n2})^{T}$
such that the values of the corresponding $I_{nj}$'s of $\hat\beta_{n1}$ and $\hat\beta_{n2}$ are 1 and 0, respectively.
Hence, the lengths of $\hat\beta_{n1}$ and $\hat\beta_{n2}$ become $\hat p_0(=\sum_jI_{nj})$ and $p -\hat p_0$.
It follows from linear algebra, we know that there exists an orthonormal matrix $O_n$, depending on the samples up to the current stage, such that ${O_n}^{T}{O_n}=I_p$ and $(\hat\beta^{T}_{n1},\hat\beta^{T}_{n2})^{T}={O_n}\hat\beta_n$.

Under this setup, we will describe our subject selection strategy and stopping criterion below.
We adopt both the D-optimality from the theory of experimental design and the concepts of uncertainty sampling in our subject selection scheme, and will separately discuss them below.

\subsection{Subject selection strategy}

\paragraph{Sequential D-criterion}
Let $A$ and $B$ be the active sample set (training data  under the current stage) and unlabeled sample pool, respectively.
For each $x^*\in B$, we compute
\begin{align}
d(x^*)={\mbox {det}} \left(\sum_{i=1}^nx_i\dot{\mu}(x_i^T\hat\beta_n)x_i^T+x^*\dot{\mu}(x^{*T}\hat\beta_n)x^{*T}\right).\nonumber
\end{align}
Ranking the set $\{d(x^*): x^*\in B\}$ in decreasing order, take the covariates from the first $\rho T_n$ largest ones as uncertainty set $U$, where $\rho$ is a pre-specified constant and $T_n$ is number of samples in $B$.

\paragraph{Uncertainty sampling strategy}
For each $x^*\in U$, we compute
\begin{align}
u(x^*)=|\mu(x^{*T}\hat\beta_n)-p_t|,\nonumber
\end{align}
where $p_t$ is a given target value.
We select the covariates $\tilde{x}$ with the minimum value in $\{u(x^*): x^*\in U\}$, then delete $ \tilde{x}$ from $B$ and add $(\tilde{y}, \tilde{x})$ in $A$, where $\tilde{y}$ is observed response value for $\tilde{x}$.

Although we use the experimental design criterion to find the promising candidate subjects  for building the classification model, there is still a major difference between the experimental design and the subject selection of active learning. In conventional design of experiments, people will construct the design points/locations first and then conduct actual observations on these particular design points.  However, in most the active learning scenarios, subjects already exist and we only use the design criteria to find these possible candidate subjects.  Hence, when the size of data set is large, how to efficiently searching the most informative ones among all subjects is an important computational issue in active learning.  Here we use the uncertainty sampling method to confine the searching range, which can help to shorten the searching time. 

\subsection{Variable selection and stopping criterion}
The relation between the logistic-type classification function and AUC are intensively discussed 
\citep[see][]{Eguchi-Copas2002}.
These results suggest us to apply sequential confidence set estimation methods to a logistic model-based active learning procedure. 
In addition to the conventional sequential estimation problems, we also merge a variable selection method to
find the effective variables with the proposed learning process, 
which will make the concluding model more compact such that we can have a good ability of model-interpretation.  

To define the stopping criterion and to identified effective variables, we will first partition matrix as follows.
Denote
$$\sum_{i=1}^nx_i\dot{\mu}(x_i^T\hat\beta_n)x_i^T
=(X_nW^{1/2})(X_nW^{1/2})^T,$$
where $\dot{\mu}(t)=\exp{(t)}/(1+\exp{(t)})^2$ and
$W=\mbox{diag}\{{\dot{\mu}(x_i^T\hat\beta_n)}, i=1,2,\ldots,n\}.$
Partitioning the matrix $(O_nX_nW^{1/2})(O_nX_nW^{1/2})^T$ according to the first $\hat p_0$ components of $O_n\hat\beta_n$
such that
\begin{align}
&(O_nX_nW^{1/2})(O_nX_nW^{1/2})^T = \nonumber \\ 
&\left( 
\begin{array}{cc}
{\Sigma_{11}(n)}_{\hat p_0\times\hat p_0} & {\Sigma_{12}(n)}_{\hat p_0\times(p_0-\hat p_0)} \\
{\Sigma_{21}(n)}_{(p-\hat p_0)\times\hat p_0}&{\Sigma_{22}(n)}_{(p-\hat p_0)\times (p-\hat p_0)}\\
\end{array}\right).
\end{align}
It implies that with simple matrix computation, 
\begin{align}\label{sigma-1}
&O_nI_n(\epsilon)\left\{(X_nW^{1/2})(X_nW^{1/2})^T\right\}^{-1}  I_n(\epsilon){O_n}^T \nonumber \\
&=\left(
\begin{array}{cc}
{\tilde{\Sigma}_{11}}^{-1}(n) & 0 \\
0 & 0 \\
\end{array}\right),
\end{align}
where
$$\tilde{\Sigma}_{11}^{-1}(n)={\Sigma^{-1}_{11}}(n)
+{\Sigma^{-1}_{11}}(n)\Sigma_{12}(n){\Sigma^{-1}_{22.1}}(n)\Sigma_{21}(n){\Sigma^{-1}_{11}}(n)$$
and
$${\Sigma^{-1}_{22.1}}(n)=\Sigma_{22}(n)-\Sigma_{21}(n) {\Sigma^{-1}_{11}}(n) \Sigma_{12}(n).$$
Let $M^{-}$ denote a general inverse matrix of $M$.
It follows that
\begin{align} \label{eq:partition}
&(Z-\hat\beta_n)^T\left[I_n(\epsilon)\left\{(X_nW^{1/2})(X_nW^{1/2})^T\right\}^{-1}I_n(\epsilon)\right]^{-}(Z-\hat\beta_n) \nonumber\\
%&=(O_nZ-O_n\hat\beta_n)^T[O_nI_n(\epsilon){O_n}^T((O_nX_nW^{1/2})(O_nX_nW^{1/2})^T)^{-1}O_nI_n(\epsilon){O_n}^T]^{-}(O_nZ-O_n\hat\beta_n)\nonumber\\
&=(Z_{n1}-\hat\beta_{n1})^T\tilde{\Sigma}_{11}(n)(Z_{n1}-\hat\beta_{n1}),
\end{align}
where $Z=(z_1,z_2,\ldots,z_p)^T\in R^p$ and $Z_{n1}$ is sub-vector of $Z$ corresponding to $\hat\beta_{n1}$.
Then Theorem \ref{Th1.2} implies that as $ n \rightarrow\infty$,
\begin{align}\label{chi2}
&(\hat\beta_n-\beta_0)^T\left[I_n(\epsilon)\left\{(X_nW^{1/2})(X_nW^{1/2})^T\right\}^{-1}I_n(\epsilon)\right]^{-}(\hat\beta_n-\beta_0)\nonumber\\
&=(\hat\beta_{n1}-\beta_{01})^T\tilde{\Sigma}_{11}(n)(\hat\beta_{n1}-\beta_{01})\rightarrow{\chi}^2(p_0) {\hbox{ in distribution}}.
\end{align}
Suppose that $d ( > 0)$ is a pre-fixed constant.
Let $S_n=(Z_{n1}-\hat\beta_{n1})^T\tilde{\Sigma}_{11}(n)(Z_{n1}-\hat\beta_{n1})$,
and $\nu_n$ be the maximum eigenvalue of $\rho(n)I_{n}(\epsilon)\{\sum_{i=1}^n x_i\dot{\mu}(x_i^T\hat\beta_n)x_i^T\}^{-1}I_{n}(\epsilon)$.
Then for a given $d>0$,
\[
R_n=\left\{Z\in R^p :\frac{S_n}{\rho(n)}\leq \frac{d^2}{\nu_n}  \mbox{  and  } z_j=0  \mbox{ for }  I_{nj}(\epsilon)=0,  1\leq j \leq p \right\}
\]
defines a confidence ellipsoid for $\beta_0$ with the length of its maximum axis no greater than $2d$.
Moreover, it follows from  Theorems \ref{Th1.1}, \ref{Th1.2} and Equation \eqref{chi2},  we have that
$\lim_{n\rightarrow\infty} P(\beta_0 \in R_n)=1-\alpha$.

Let $\hat p_0(n)$ be the estimate of $p_0$ based on the first $n$ observations as defined in Theorem \ref{Th1.1}.
Because the true $p_0$ in \eqref{chi2} is unknown, we replace it with a strongly consistent estimate $\hat p_0$.
Let $C_n = \{(y_i,x_i):~i=1,\cdots,n\}$ be the set of the first $n$ observations, and for given $C_n$,
let $a_n^2 \in R$ be a constant satisfying the conditional probability $P(\chi_{\hat p_0(n)}^2\leq a_n^2 |~ C_n)=1-\alpha$ for a given $\alpha$.
Thus, for a given observed samples $C_n$, $\hat p_0(n)$ is a constant, and $a_n^2$ is an $1-\alpha$ quantile of the chi-square distribution with $\hat p_0(n)$ degrees of freedom.
Now, suppose $\{(y_1,x_1),\cdots,(y_{n_0},x_{n_0})\}$ is a set of subjects in the beginning with $n_0 (\geq p)$. Then a stopping time {$N_d$} is defined as follows:
\begin{align}\label{Ndef}
N=N_d\equiv\inf\left\{n:~n\geq n_0~~\mbox{and}~~\nu_n\leq \rho(n) d^2/a_n^2\right\},
\end{align}
where $a_n^2$ and $d>0$ are two constants defined before.
Equation \eqref{Ndef} means that we will stop recruiting new samples into our training set once the maximum eigenvalue $\nu_n$ satisfies the inequality in \eqref{Ndef}.
Replacing the non-random sample size $n$ in \eqref{eq:partition} with the newly defined stopping time $N=N_d$, 
we define $S_N=(Z_{N1}-\hat\beta_{N1})^T\tilde{\Sigma}_{11}(N)(Z_{N1}-\hat\beta_{N1})$.
Similarly, we have
\begin{align}\label{Rn}
R_N\!\! =\!\left\{ Z \in R^p\!\!:~~\!\!\frac{S_N}{\rho(N)}\leq
\frac{d^2}{\nu_N}\! \mbox{~~ and~~ }\! z_j\!=0 \!\mbox{~~ for~~ }\!
I_{Nj}(\epsilon)\!=\!0, 1\leq j\leq p\right\},
\end{align}
which is a sequential fixed size confidence ellipsoid for $\beta_0$.

The following theorem says that using both uncertainty sampling and the D-optimal design method to selection new training subjects, sequentially, the proposed an active learning procedure will have the following properties. (The proof of Theorem \ref{Th1.4} will be given in Appendix.)
\begin{theorem}\label{Th1.4}
Assume that $\{(x_i, y_i), i \geq 1\}$ follows the logistic regression model
(\ref{model}), Conditions (A1) -- (A3) are satisfied and $\sup_i||x_i||<\infty$ almost surely. Let $N$ be
defined as in (\ref{Ndef}). Then
\begin{itemize}
\item[] \text
{(i)} $\lim_{d\rightarrow 0}d^2N/(a^2\nu)=1 ~~\text{\rm almost surely}$, %\label{sconclu1}
\item[] \text
{(ii)} $\lim_{d\rightarrow 0} P(\beta_0 \in R_N)=1-\alpha$, %\label{sconclu2}
\item[] \text
{(iii)} $\lim_{d\rightarrow   0}d^2E(N)/(a^2\nu)=1$, %\nonumber %\label{sconclu3}
\item[] \text
{(iv)} $\lim_{d\rightarrow 0}\hat p_0(N) = p_0$ almost surely, and $\lim_{d\rightarrow 0}E(\hat p_0(N))=p_0$,
\end{itemize}
where $\nu$ is the maximum eigenvalue of matrix ${I_0}\Sigma^{-1} {I_0}$.
\end{theorem}

Note that under active learning scenarios described here, when we recruit new subjects to join our training set, 
we do not know their label information. 
Their label information will only be revealed after being selected, 
and we estimate the regression coefficient vector with the selected subjects only.
Because we sequentially find the effective variables using the current training subjects, 
the degree of freedom of the asymptotic $\chi^2$ distribution is also data-dependent,
which makes this sequential estimation procedure here different from the conventional ones.

It is clear from the definition that the stopping time $N$ will go to infinity as $d$ goes to 0.
Theorem \ref{Th1.4} (ii) and (iii) say that if $d$ goes to 0, then
the coverage probability of $R_N$ asymptotically equals to the nominated value $1-\alpha$ 
and  the expected sample size of the sequential procedure approaches to the best (unknown) sample size.
In \cite{Chow1965}, they called these two properties as asymptotic consistency and asymptotic efficiency 
of sequential estimation methods, respectively, 
In addition, Theorem \ref{Th1.4} (iv) states that $\hat p_0(N)$ almost surely converges to
number of the effect parameters under the proposed sequential procedure.
Hence, Theorem \ref{Th1.3} and Theorem \ref{Th1.4} (iv) together implies that the effective variables are eventually identified.
In practice, choices of constant $d$ depends on the application needs and many other factors.
A smaller $d$ means that we required more accurate/precise estimate, hence a larger sample size is usually required.

\subsection{Stopping criterion and area under ROC curve}
In \cite{Eguchi-Copas2002} , they showed that when a logistic model is correct, then its classification function will reach the theoretical maximum AUC, asymptotically.  Here, we will study that whether the proposed procedure have a satisfactory AUC with the proposed stopping criterion.

Let $\hat\theta_n$ be the angle between $\hat\beta_n$ and $\beta_0$, and
AUC$_{\hat\beta_n}$ and AUC$_{\beta_0}$ be the AUCs of with respect to $\hat\beta_n$ and $\beta_0$.
Because AUC is scale-invariant, to show that AUC$_{\hat\beta_N}$ converges to AUC$_{\beta_0}$,
it is sufficient to show that $\hat\theta_N$ converges to 0.
We know that from the definition, $N_d$ goes to infinity as $d$ goes to 0 with probability one.
Since Theorem \ref{Th1.1} and \ref{Th1.2} together imply that ${\hat{\beta}_{N}} \rightarrow \beta_0$ almost surely as $d$ goes to 0, they also imply that $\hat\theta_N$ converges to 0 almost surely.
Thus, we have a corollary below, which shows that the empirical AUC will also reach its theoretical optimal with the proposed procedure.
\begin{corollary}\label{cor}
Let $\hat\theta_n$ be the angle between $\hat\beta_n$ and $\beta_0$, then
under the assumptions of Theorem \ref{Th1.4}, $\hat\theta_N \rightarrow 0$ as $d$ goes to 0.
In addition,  AUC$_{\hat\beta_N}$ $\rightarrow$ AUC$_{\beta_0}$ almost surely
as $d$ goes to 0.
\end{corollary}
(Proof follows from simple algebra operations and is given in Appendix.)

\section{Numerical results}
We report the numerical results of the proposed ASE-based active learning procedure using some synthesized data sets, and compare with the results obtained from MLE-based active learning methods.  
In addition, we use the credit card fraud detection  and the MAGIC gamma telescope data sets obtained from the Internet  for demonstration purposes.  We will describe these two data sets later.

\subsection{Synthesized Data}
We generate $30000$ synthesized data based on a logistic regression model stated in \eqref{model}
with the coefficients $\beta_1=-1$, $\beta_2=1$ and $\beta_3=\beta_4=0$ (i.e. $p=4$),
with  $x=(x_1,x_2,x_3)^T$ generated from a multivariate normal distribution, $N(0, I_3)$,  where $I_3$ is a  $3 \times 3$ identity matrix. In addition to the uncertainty sampling,  an optional clustering algorithm, using only the covariate vectors, is used to partition the training data in order to reduce the searching time.

The results in Tables \ref{tab2} and \ref{tab3} are based on \textcolor{black}{1000 simulations}, and
Table 3 also lists estimate of number $p_0$ of nonzero parameters for the proposed procedure.  
The results of these two tables show that when classification performances of two methods are similar,  the proposed procedure will stop earlier (use less training subjects), have a shorter computational times and the concluding model is more compact model (less variables used) than that of the MLE-based procedure.
In addition, it shows that when $d$ becomes small, the estimates of regression parameters and $p_0$ will converge to their corresponding true values, and standard variances of these estimates will also decrease.

\begin{table}[!h]
\selectfont
\begin{center}
\caption {Estimates of stopping time and prediction performance.}\label{tab2}
\vskip 10pt
\begin{tabular}{ccrcccc}
\hline
%&&\multicolumn{3}{c}{d}\\
%\cline{3-5}
 d& Method & N & $\kappa$ & time & ACC & AUC \\
\hline
0.5 &ASE &  244.957(91.271)&1.035&0.322(0.146)&0.657(0.046)&0.658(0.044) \\
&MLE &  1008.418(570.487)&0.920&2.562(2.297)&0.674(0.033)&0.674(0.033) \\
0.4 &ASE &  387.794(117.275)&1.019&0.596(0.228)&0.662(0.043)&0.663(0.041) \\
 &MLE &  1317.388(545.162)&0.863&3.646(2.375)&0.674(0.032)&0.675(0.033) \\
0.3 &ASE &  727.966(144.985)&1.009&1.614(0.474)&0.669(0.037)&0.670(0.036) \\
 &MLE &  1731.712(371.836)&0.723&5.148(2.022)&0.675(0.032)&0.675(0.032) \\
\hline
%\multicolumn{10}{l}{$^*$ Number of non-zero parameters estimated as 0. }\\
%\multicolumn{10}{l}{$^*$ Number of zero parameters estimated as 0. }
\end{tabular}
\end{center}
\end{table}

\begin{table}[!h]
\selectfont
\begin{center}
\caption {Estimates of regression parameters.}\label{tab3}
\vskip 10pt
\begin{tabular}{c cc c c c c}
\hline
%&&\multicolumn{3}{c}{Balance}&&&\multicolumn{3}{c}{unbalace}\\
%\cline{3-5}\cline{8-10}
 & d &$\hat p_0$ &$\beta_1=-1$ & $\beta_2=1$ & $\beta_3=0$ &$\beta_4=0$ \\
\hline
ASE &0.5 & 1.804(0.400)&-0.835(0.442) & 0.972(0.185)&-0.001(0.025)&0.001(0.029) \\
 &0.4 & 1.862(0.351)&-0.876(0.378) & 0.978(0.141)&0(0)&0(0.015) \\
 &0.3 & 1.940(0.242)&-0.941(0.263) & 0.989(0.100)&0(0)&0(0.006) \\

MLE &0.5 & -&-1.005(0.117) & 1.008(0.087)&-0.001(0.056)&0.002(0.054) \\
 &0.4 & -&-1.003(0.099) & 1.003(0.072)&0(0.05)&0.002(0.049) \\
 &0.3 & -&-1.000(0.081) & 1.000(0.062)&0(0.048)&0.002(0.046) \\
\hline
%\multicolumn{10}{l}{$^*$ Number of non-zero parameters estimated as 0. }\\
%\multicolumn{10}{l}{$^*$ Number of zero parameters estimated as 0. }
\end{tabular}
\end{center}
Note that there is no variable selection in the MLE-based procedure, hence there is no estimate of $p_0$ in these cases.
\end{table}

\subsection{Real Examples}

%--------
\subsubsection*{Credit Card Fraud Detection Data set}
The credit card fraud detection is an anonymized data set obtained from  a machine learning competition website {\it Kaggle}, 
which is a platform for data science competitions. Please refer to their website \hbox{\em www.kaggle.com/host} 
for the further details.

In this data set, it consists of transactions occurred in two days of September 2013 by European cardholders.
There are 492 frauds out of 284,807 transactions; i.e there are only 0.172\% of all transactions are frauds. 
For this data set, we will refer to these fraud transactions as positive cases.
Due to confidentiality issues, they did not offer the original features and the detailed background information about this data set, and only numerical variables resulting from a PCA transformation with 28 principal components are available.
The feature ``Amount'' is the transaction amount, which is not included in the PCA transformation,
and ``Class'' is the response variable, which takes value 1 in case of fraud and 0 otherwise.
We apply our procedure to the credit card fraud detection data set, and use the first 3 and last 2 components in our analysis.  
We select these 5 PCA components as covariates, since we want to show both the classification performance and variable identification ability.  The variables names Var1, Var2 and Var3 denote the first 3 PCA components and Var27 and Var28 denote the last two components.
In each simulation run, we randomly select 400 fraud cases, and 1600 from the regular cases, so the total size of the training set is 2000.
Thus, there are 282807 regular cases and 92 fraud cases in our testing set.
Based on this setup, we expect that our method can successfully find the first three effective components, and
produce a satisfactory prediction results as well. (See the top three plots of Figure \ref{fig:real-est}.)
Because the ratio of the sizes of the cases to the non-cases is small,  using AUC as the performance measure is also recommended (see also Kaggle website). 
Here, we summarize both averages of the accuracy and AUC based on 1000 runs, and we also report 
the variable selection results and their corresponding coefficient estimates in addition.
%--------

\subsubsection*{MAGIC Gamma Telescope Data Set}
We get this MAGIC Gamma Telescope Data Set from the UC Irvine Machine Learning Repository (\hbox{\em archive.ics.uci.edu/ml}), and according to their description, this data set is used to simulate the registration of high energy gamma particles in a ground-based atmospheric Cherenkov gamma telescope using the imaging technique.

\begin{figure}[h]
    \centering
    Credit Card Fraud\\
         \includegraphics[width=0.3\textwidth]{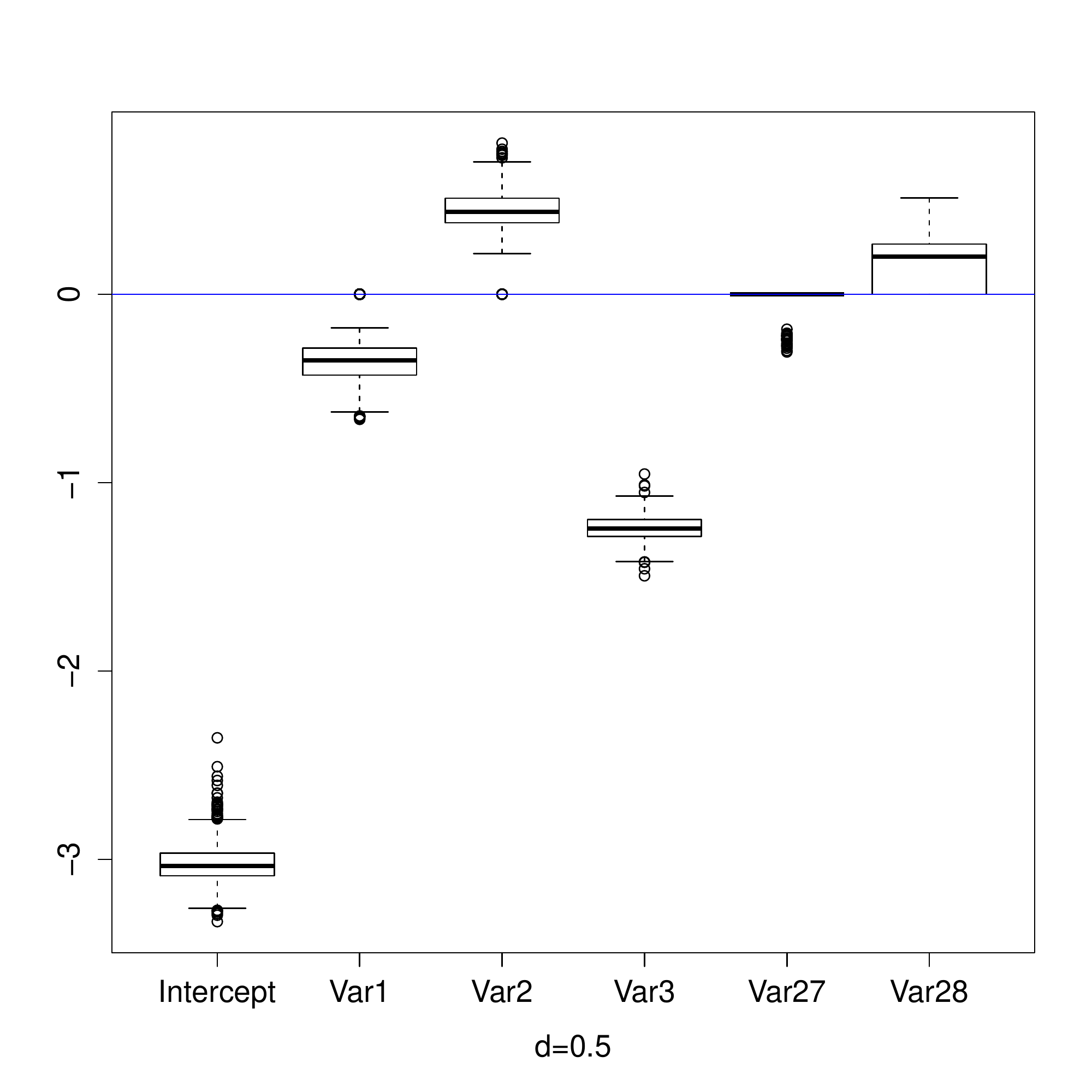} 
          \includegraphics[width=0.3\textwidth]{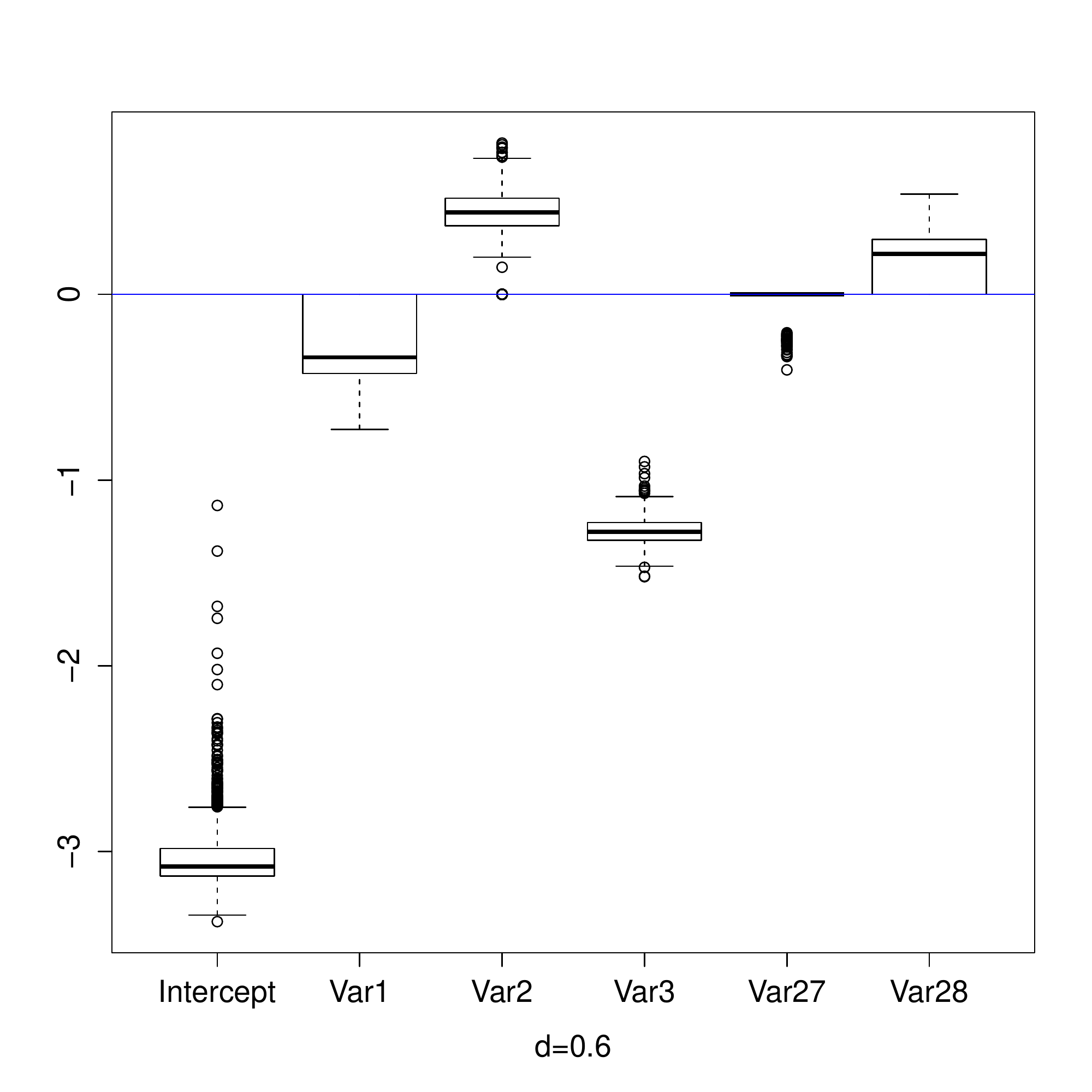} 
         \includegraphics[width=0.3\textwidth]{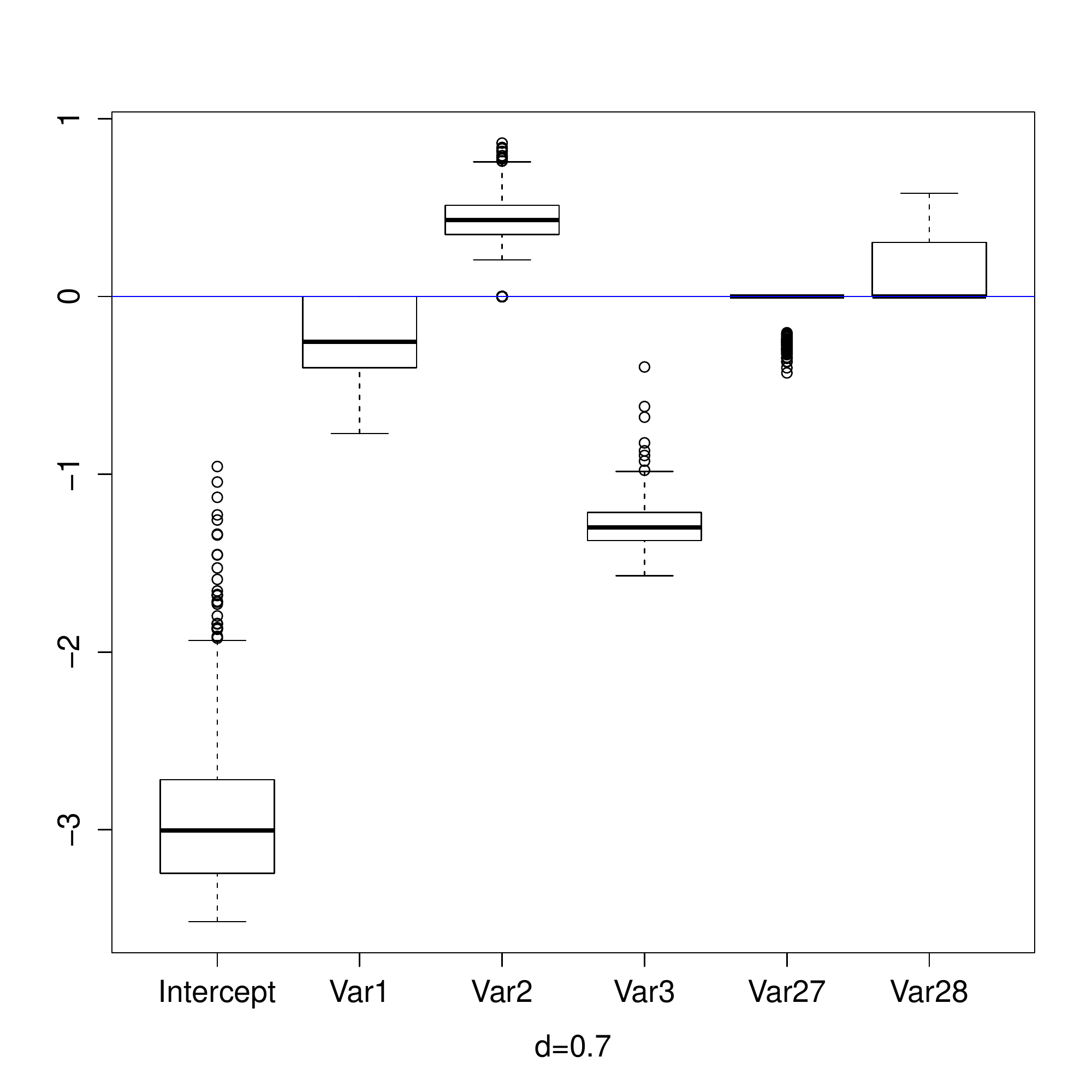}   \\
    Magic Gamma\\
         \includegraphics[width=0.3\textwidth]{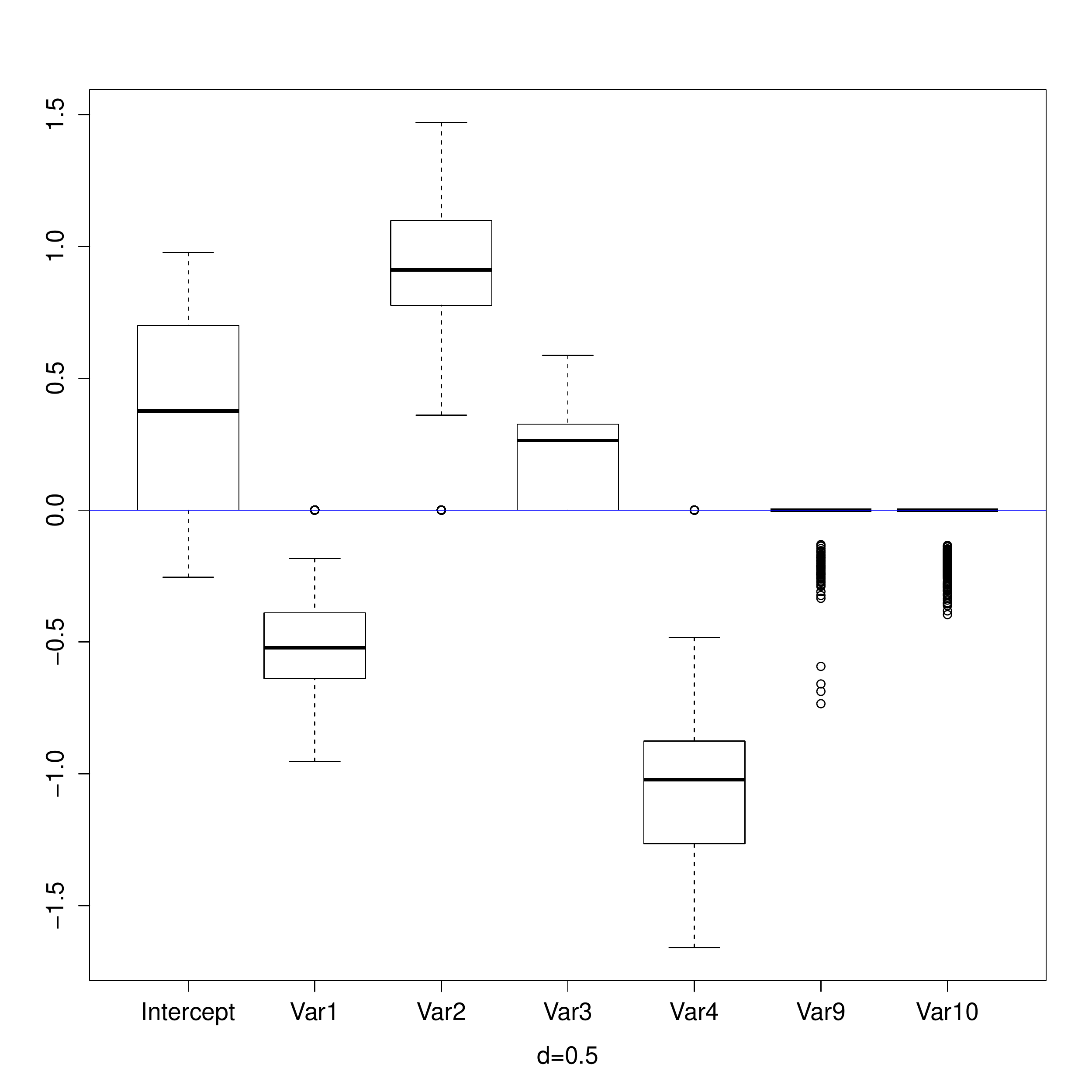} 
          \includegraphics[width=0.3\textwidth]{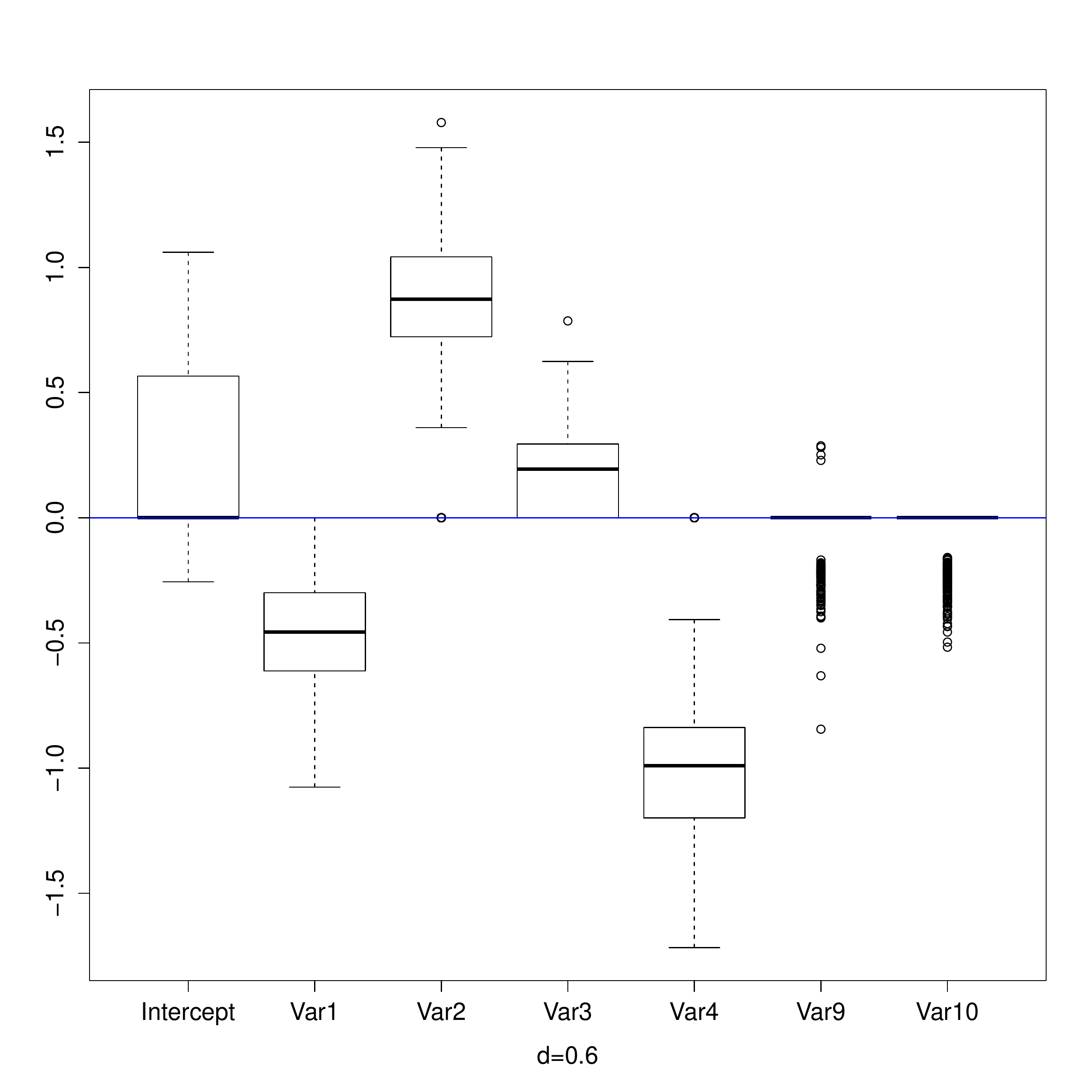} 
         \includegraphics[width=0.3\textwidth]{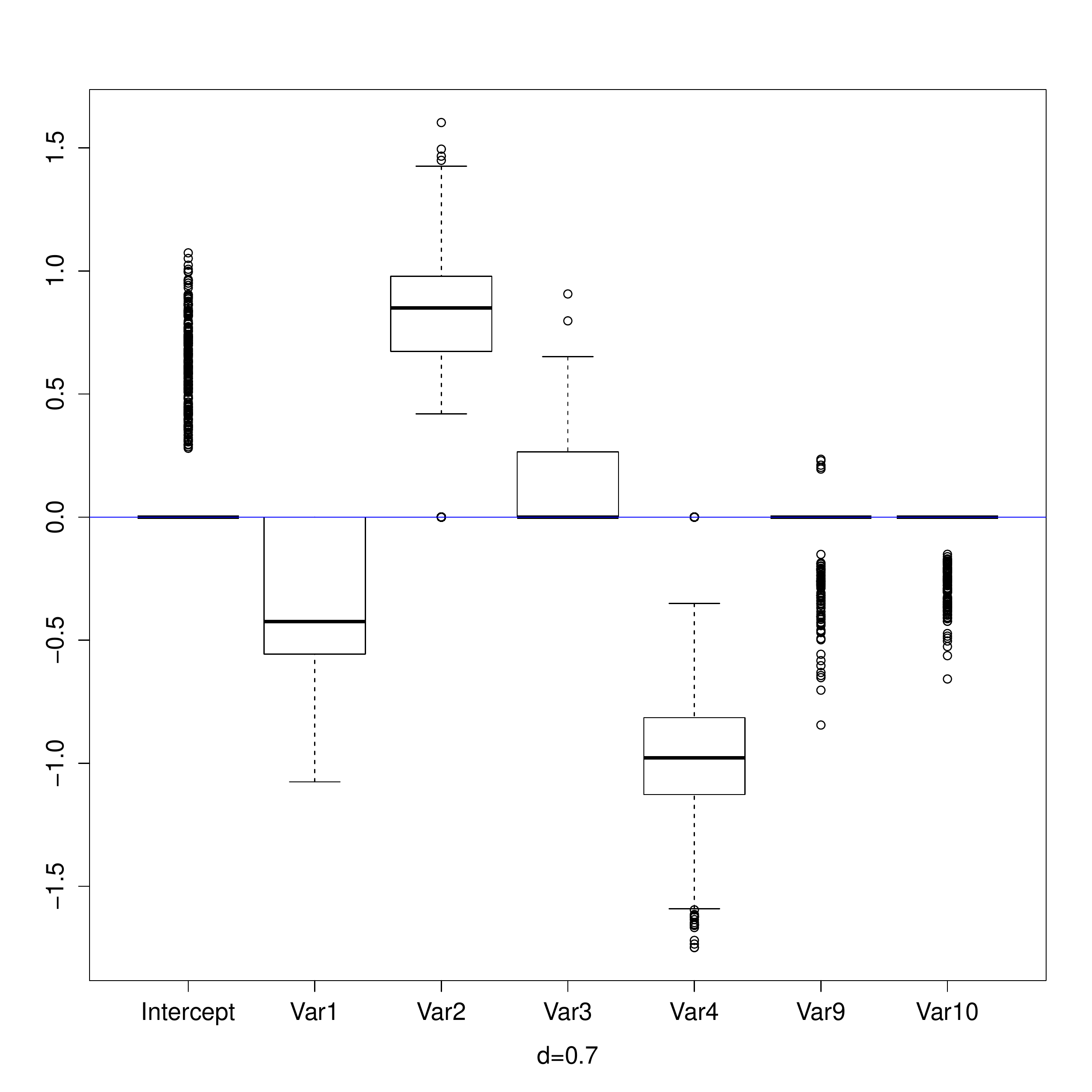} 
         
             \caption{Box-plots of Coefficient Estimates}\label{fig:real-est}

 \end{figure}

There are 10 continuous real-valued variables and a class variable.  
The total sample size is 19020. 
Among them there are 12332  signal (gamma) and 6688 background (hadron) samples, 
and we will refer to these subjects with gamma signal as positive cases, and the other 6688 subjects as negative cases.
We conduce a similar PCA transformation as the previous example, and use the first 4 and last 2 PCA components as our covariates.
Hence, our procedure should only select the first 4 variables. (See the bottom three plots of Figure \ref{fig:real-est}.)
For each simulation run, we randomly select 20\% of subjects as our training set.  In order to keep the positive to negative ratio, there are 2466 positive subjects and 1338 negative subjects in the training set.  We repeat this scheme 1000 runs.
Our goal is to discriminate statistically those caused by primary gammas (signal) from the images of hadronic showers initiated by cosmic rays in the upper atmosphere (background).  The detailed explanation of this data set and its physical background, can be found in the UC Irvine Machine Learning Repository and the original owner's website (\hbox{\em www.magic.mppmu.mpg.de}).

Table \ref{tab:real} shows the average number of the training subjects (stopping time) used, and the prediction performance (both accuracy and AUC), when we apply the ASE-based active learning procedure to these two data sets with 1000 runs and different $d$s.  
Figure \ref{fig:real-est} shows the box-plots of coefficient estimate based on 1000 runs for each case, separately.
For the Credit Card Fraud data set, the first 3 variables (PCA components) are successfully identified; especially with the case of$d=0.5$, the estimate of the intercept term of the model reflects the imbalanced ratio of the sizes of the cases to the controls. 

Note that in simulation study, we already know that the required training subject size of the MLE-based active learning procedure is 2 to 4 times of that of the corresponding ASE-based method (see Table \ref{tab2}). Because it will take too much time for the MLE-based methods, we only apply the ASE-based method to these two real data sets.

\begin{table}[h!]
    \small
            \caption {Average stopping time and prediction performance of ASE-based active learning procedures}
    \centering
    \label{tab:real}
    \begin{tabular}{*{8}{c}}
    \hline
    Data & d & N & Time & ACC & AUC \\
    \hline
    Credit & 0.5 &  1446.18 (208.90) &  18018.49 (5981.34) & 0.9801 (0.0027) & 0.8378 (0.0227) \\
    Card  & 0.6 &  1024.39 (215.67) &  11696.80 (4571.01) & 0.9797 (0.0040) & 0.8378 (0.0229) \\
    & 0.7 &  740.81 (262.26) &  7552.82 (4281.15) & 0.9780 (0.0068) & 0.8384 (0.0231) \\
    \hline
    Magic & 0.5 &  719.48 (347.07) &  2699.02 (2622.26) & 0.7994 (0.0241) & 0.6462 (0.0304) \\
      Gamma & 0.6 &  441.54 (246.15) &  1824.43 (2021.63) & 0.8107 (0.0241) & 0.6391 (0.0335) \\
     & 0.7 &  293.53 (158.71) &  697.18 (888.25) & 0.8188 (0.0210) & 0.6322 (0.0351) \\
    \hline
     \end{tabular}
 \end{table}

\iffalse
%Figure - Boxplot of stopping time
\begin{figure}
    \centering
         \includegraphics[width=0.45\textwidth]{Rplot_N.pdf} 
         \quad
         \includegraphics[width=0.45\textwidth]{Magic_Rplot_N.pdf} 
    \label{fig:test}
       \caption[]{Box-plots of Stopping Times}\label{fig:real-stop}
\end{figure}
\fi

%Figure - Estimated coefficients Boxplot

%----------------------

\section{Conclusion}
In this paper, we propose a procedure for building binary classification models when the complete label information is not available in the beginning of the training stage. 
To this end, we apply the idea of active learning procedure to a logistic model  such that the proposed procedure can simultaneously select the most informative subjects for training and find the effective variables for this classification model.  

In an active learning procedure, we continuously select new observations according to some predefined selection criteria.
until a predetermined stopping criterion is fulfilled.
Here we use the information obtained from analyzing the current observations 
to select the most informative observations from a given training data set in order to shorten the learning course,
and this selection scheme is different from the conventional sequential analysis.
We adopt an adaptive selection procedure in the proposed algorithm,
and then treat the dependent-observation situations as a logistic model with adaptive covariates.

Because we use a parametric logistic model in our active learning procedure, 
we are able to use the methods of experimental design to find the most informative subjects from the given training data set.
The criterion used here is like that of the conventional sequential optimal experimental design, 
however, in an active learning process we just use this criterion  to search the potential observations from the existing data set.
Thus, the key step will be how to search the most informative subjects in the given data set.
We do not have to know exactly where the optimal design points are as in the traditional optimal experimental design problems.
Instead, we will select the observations that are ``close'' enough to the theoretical optimal design points.

We take the advantages of stochastic approximations and optimal design,
and aims to search the next unlabeled data point(s) from huge pool of subjects with the aid of the uncertainty sampling.
The uncertainty sampling strategy is vague concept and here we only use this idea to confine our searching range of new samples to shorten the searching time. 
The asymptotic results presented here only assumed the selected observations satisfying some regularity conditions and do not depend on any specific design criterion,
hence the other selection criteria can also be used in our procedure.  
Here, we use the D-optimality criterion and the uncertainty sampling strategy together.
Depending on the learning target, we can use other indexes to replace this criterion.
Moreover, the relations between the optimal experimental design and the method of uncertainty is not studied here,
which will be an interesting future study problem.

\color{black}
\section*{Appendix A.}%Locate effective variables}
\noindent{\bf{Proof of Theorem \ref{Th1.1}}}:
Let $\eta(n)=L$.
Firstly we consider non-zero component $\beta_{0j}\neq 0$ for some $j$,
$1\leq j \leq p$.  For any $\eta > 0$,  we have
\begin{align}%\label{ieq1}
&P(\sup_{n\leq k}|I_{kj}(\epsilon)-1|>\eta)\nonumber\\
&= P(\sup_{n\leq k}|I(\sqrt{\eta(k)}\lambda
|\tilde{\beta}_{kj}|^{-\gamma}< \epsilon)-1|>\eta)\nonumber\\
&= P(\sup_{n\leq k}|I(\sqrt{\eta(k)}\lambda
|\tilde{\beta}_{kj}|^{-\gamma}< \epsilon)-1|>\eta, \sup_{n\leq
k}\sqrt{\eta(k)}\lambda
|\tilde{\beta}_{kj}|^{-\gamma}< \epsilon)\nonumber\\
& \hspace{0.3cm} +P(\sup_{n\leq k}|I(\sqrt{\eta(k)}\lambda
|\tilde{\beta}_{kj}|^{-\gamma}< \epsilon)-1|>\eta, \sup_{n\leq
k}\sqrt{\eta(k)}\lambda
|\tilde{\beta}_{kj}|^{-\gamma}\geq \epsilon)\nonumber\\
&\leq P(\sup_{n\leq
k}\sqrt{\eta(k)}\lambda|\tilde{\beta}_{kj}|^{-\gamma}\geq \epsilon).\nonumber
\end{align}
From \cite{chenhuying1999} or \cite{Chang2001}, we already know that $\tilde{\beta}_{n}-\beta_0\rightarrow 0$ almost surely, as
$n\rightarrow \infty$.
This implies that $P(\sup_{n\leq
k}|\tilde{\beta}_{kj}-\beta_{0j}|>\eta)< \eta$ for large enough $n$.
Hence,
\begin{align}%\label{ieq2}
P(\sup_{n\leq k}\sqrt{\eta(k)}\lambda|\tilde{\beta}_{kj}|^{-\gamma}\geq
\epsilon) &\leq P(\sup_{n\leq
k}\sqrt{\eta(k)}\lambda|\tilde{\beta}_{kj}|^{-\gamma}\geq \epsilon,
\sup_{n\leq k}|\tilde{\beta}_{kj}-\beta_{0j}|>\eta)\nonumber\\
& \hspace{0.3cm}+ P(\sup_{n\leq
k}\sqrt{\eta(k)}\lambda|\tilde{\beta}_{kj}|^{-\gamma}\geq \epsilon,
\sup_{n\leq k}|\tilde{\beta}_{kj}-\beta_{0j}|\leq \eta)\nonumber\\
&\leq \eta+P(\sup_{n\leq k}\sqrt{\eta(k)}\lambda(|\beta_{0j}|-\eta)^{-\gamma}\geq \epsilon)\nonumber\\
&=\eta+P(\sup_{n\leq k}\sqrt{\eta(k)}\lambda\geq c\epsilon).\nonumber
\end{align}
Since $n^{1/2}\lambda \rightarrow 0$ by definition of $\lambda$, we
have
\begin{align}\label{b0jneq0}
P(\sup_{n\leq k}|I_{{ k}j}(\epsilon)-1|>\eta) \leq 2\eta .%\rightarrow 0,\mbox{ as } n\rightarrow \infty.
\end{align}

If $\beta_{0j}= 0$ for some $j$, $1\leq j \leq p$, then
\begin{align}
&P(\sup_{n\leq k}I_{kj}(\epsilon)>\eta)=P(\sup_{n\leq
k}I(\sqrt{\eta(k)}\lambda
|\tilde{\beta}_{kj}|^{-\gamma}< \epsilon)>\eta)\nonumber\\
&=P(\sup_{n\leq k}I(\sqrt{\eta(k)}\lambda |\tilde{\beta}_{kj}|^{-\gamma}<
\epsilon)>\eta, \inf_{n\leq k}\sqrt{\eta(k)}\lambda
|\tilde{\beta}_{kj}|^{-\gamma}< \epsilon)\nonumber\\
& \hbox{  }+P(\sup_{n\leq k}I(\sqrt{\eta(k)}\lambda
|\tilde{\beta}_{kj}|^{-\gamma}< \epsilon)>\eta, \inf_{n\leq
k}\sqrt{\eta(k)}\lambda
|\tilde{\beta}_{kj}|^{-\gamma}\geq \epsilon)\nonumber\\
&\leq P(\inf_{n\leq
k}\sqrt{\eta(k)}\lambda|\tilde{\beta}_{kj}|^{-\gamma}< \epsilon).\nonumber
\end{align}
In addition, we also have
$0<\delta<1/2$, $\eta(k)^{\delta}\tilde{\beta}_{kj}\rightarrow 0,$ almost surely as
$n\rightarrow \infty$ \citep{chenhuying1999} . Therefore,
\begin{align}%\label{ieq4}
P(\sup_{n\leq k}I_{kj}(\epsilon)>\eta)
&\leq P(\inf_{n\leq k}\sqrt{\eta(k)}\lambda|\tilde{\beta}_{kj}|^{-\gamma}< \epsilon)\nonumber\\
&= P(\inf_{n\leq k}\eta(k)^{(1+2\gamma\delta)/2}\lambda|\eta(k)^{\delta}\tilde{\beta}_{kj}|^{-\gamma}< \epsilon)\nonumber\\
&\leq\eta+P(\inf_{n\leq k}\eta(k)^{(1+2\gamma\delta)/2}\lambda<
c\epsilon).\nonumber
\end{align}
Due to
$\eta(n)^{(1+2\gamma\delta)/2}\lambda\rightarrow \infty$, we show
\begin{align}\label{b0jeq0}
P(\sup_{n\leq k}I_{kj}(\epsilon)>\eta) \leq 2\eta. %\rightarrow 0, \mbox{ as } n\rightarrow \infty.
\end{align}
Let $\eta$ be arbitrarily small, from
(\ref{b0jneq0}) and (\ref{b0jeq0}), we have that $I_{nj}(\epsilon)\rightarrow
I(\beta_{0j}\neq 0)$,almost surely as $n\rightarrow \infty$, and
$\lim_{n\rightarrow\infty}\hat{p}_0=p_0$, almost surely.

By the definition of $\hat{p}_0$, we know that
\begin{align}\label{Epieq0}
E(\hat{p}_0)=\sum_{i=1}^{p}E(I_{nj}(\epsilon))=\sum_{i=1}^{p}
P(\sqrt{\eta(n)}\lambda|\tilde{\beta}_{nj}|^{-\gamma}< \epsilon).
\end{align}
Since  $|I_{nj}(\epsilon)|\leq 1$ for all $j$, it follows from the Dominated Convergence Theorem,
that $\lim_{n\rightarrow\infty}E(\hat{p}_0)=p_0$. Thence,
the proof of Theorem \ref{Th1.1} is compeleted. \qed 

\iffalse
\begin{theorem}\label{Th1.2}
Suppose the assumptions of Theorem \ref{Th1.1} are satisfied.
%\textcolor{red}
Then with probability one, (i) $\hat\beta_n \rightarrow \beta_0$  and (ii)
$\|\hat{\beta}_n-\beta_0\|=O(\rho(n)^{-1/2})$ as $n \rightarrow \infty$.
(iii) If, in addition, Condition (A3) is also satisfied, then  for { any small
$\epsilon >0$},
\begin{align}%\label{sasynorm}
\rho(n)^{1/2}(\hat{\beta}_n-\beta_0)\rightarrow
N(0,\sigma^{2}{I_0}\Sigma^{-1} {I_0}) ~~\text{\rm in distribution
as } n\rightarrow \infty,\nonumber
\end{align}
where  ${I_0} = \mbox{diag}\{I(\beta_{01}\neq
0),\cdots,I(\beta_{0p}\neq 0)\}$ is a $p\times p$ diagonal matrix.
\end{theorem}
\fi

\vspace{10pt}
\noindent{\bf{Proof of Theorem \ref{Th1.2}}}:\\
By the definition of $\hat{\beta}_k$, it easily shows that
for any given $\xi>0$,
\begin{align}\label{supineq1}
P(\sup_{k{\geq} n}||\hat{\beta}_k-{I_0}\beta_0||>\xi)\leq & P(\sup_{k{\geq} n}||\tilde{\beta}_k-\beta_0||>\frac{\xi}{2}) \nonumber \\
&+P(\sup_{k{\geq} n}||\beta_0||\cdot
||I_{{k}}(\epsilon)-{I_0}||>\frac{\xi}{2}).
\end{align}
From (\ref{supineq1}) and Theorem \ref{Th1.1},
{$\hat{\beta}_n$} is a strong consistence estimator of $\beta_0$.
Since we already know that with probability one,
\begin{eqnarray}\label{lsrate}
||\tilde{\beta}_n-\beta_0||=O(L^{-1/2}).
\end{eqnarray}
Hence, using the triangle inequality, we have
\begin{eqnarray}\label{trin}
||\hat{\beta}_n-\beta_0||\leq
||I_{{n}}(\epsilon)(\tilde{\beta}_n-\beta_0)||+
||(I_{{n}}(\epsilon)-{I_0})\beta_0||.
\end{eqnarray}
It follows from (\ref{Irate}), (\ref{lsrate}) and (\ref{trin}), that
(ii) holds.
Moreover, because 
\begin{align}%\label{sasynorm}
\rho(n)^{1/2}(\tilde{\beta}_n-\beta_0)\rightarrow
N(0,\Sigma^{-1} ) ~~\text{\rm in distribution
as } n\rightarrow \infty, \nonumber
\end{align}
Theorem \ref{Th1.2} (iii) follows from the definition of $\hat{\beta}_n$ and the Slutsky's theorem.
 So, Theorem \ref{Th1.2} holds. \qed
%\vskip 10pt

 %
%It is known that $P(\pmb\beta_T \in R_T) \rightarrow 1-\alpha$ as $d \rightarrow 0$. Moreover,  by definition, the  maximum axis of $R_n$ is no greater than $2d$.  Therefore, with simple vector algebra, it is shown that as ${d\rightarrow 0}$, with probability no less than $1 - \alpha$, that
%\[
%\cos(\hat\theta) = \frac{ <\hat\ell, \ell> }{{\|\hat\ell\| \| \ell\|}} \geq \left(\frac{\|\ell\| - d}{\| \ell \| + d}\right)^2.
%\]
%It implies that with probability no less than $1-\alpha$,
%\[
%   0 \leq \hat\theta \leq \cos^{-1}\left(\left(\frac{\|\ell\| - d}{\| \ell \| + d}\right)^2\right).
%\]
%\vskip 10pt
\noindent{\bf{Proof of Theorem \ref{Th1.4}}}:\\
Under D-optimality, we choose the new samples such that the determinant of information matrix is maximized, which implies the minimum eigenvalue $\nu_{min}(n)$ has order of $n$.
Uncertainty sampling step selects one sample from the chosen sample set of D-optimality. It implies that
$\nu_{min}(n)/n$ is larger than 0 for all $n$. Therefore, the proof of Theorem \ref{Th1.4} follows the similar arguments
to those of \citet[][Theorem 8]{wangchang13} in  and Lemma 3.3, Theorems 3.1 and 3.2 in \cite{Chang2001}.  Hence, the details will be omitted here.

\vspace{10pt}
\noindent{\bf{Proof of Corollary \ref{cor}}}:\\
From the definition of stopping time, we know that $N$ goes to infinity  as $d$ goes to 0 with probability one.
It has been shown that ${\hat{\pmb\beta}_{N}} \rightarrow \pmb\beta_0$ almost surely.
This implies that $\hbox{AUC}_{\pmb{\hat\beta}_{N}}$ eventually converges to $\hbox{AUC}_{\pmb\beta_0}$ as ${d\rightarrow 0}$.
Similarly, it is known that $P(\pmb\beta_0 \in R_N) \rightarrow 1-\alpha$ as $d \rightarrow 0$.
Moreover,  by definition, the  maximum axis of $R_n$ is no greater than $2d$.
Let $\hat{\pmb\ell}$ denote the estimated direction  $\pmb{\hat\beta}_{N}$.
Then, with simple vector algebra, it is shown that as ${d\rightarrow 0}$, with probability no less than $1 - \alpha$, that
\[
\cos(\hat\theta) = \frac{ <\hat{\pmb\beta}_N, \pmb\beta_0> }{{\|\hat{\pmb\beta}_N\| \| \pmb\beta_0\|}} \geq \left(\frac{\|\pmb\beta_0\| - d}{\| \pmb\beta_0 \| + d}\right)^2.
\]
It is clear that if $d$ goes to 0, then $\cos(\hat\theta)$ goes to 1, which implies that  $\hat\theta$ converges to 0.
It implies that with probability no less than $1-\alpha$, if $d$ is small enough, then
\[
   0 \leq \hat\theta \leq \cos^{-1}\left(\left(\frac{\|\pmb\beta_0\| - d}{\| \pmb\beta_0 \| + d}\right)^2\right),
\]

\end{document}